\documentclass[twocolumn, switch]{article}
\usepackage{preprint}
\usepackage{hyperref}
\usepackage{amsmath}
\usepackage{booktabs}
\usepackage{mathtools}
\usepackage[numbers,square]{natbib}

\hypersetup{
colorlinks=true, 
linkcolor=[rgb]{0.5, 0.0, 0.0},
citecolor=[rgb]{0.2, 0.4, 0.9}, 
urlcolor=blue, 
anchorcolor=red
}

\usepackage[utf8]{inputenc}	
\usepackage[T1]{fontenc}
\usepackage{xcolor}
\usepackage{lineno}					
\usepackage{tikz} 					

\usepackage{newfloat}
\DeclareFloatingEnvironment[name={Supplementary Figure}]{suppfigure}
\usepackage{sidecap}
\sidecaptionvpos{figure}{c}

\usepackage{titlesec}
\titlespacing\section{0pt}{12pt plus 3pt minus 3pt}{1pt plus 1pt minus 1pt}
\titlespacing\subsection{0pt}{10pt plus 3pt minus 3pt}{1pt plus 1pt minus 1pt}
\titlespacing\subsubsection{0pt}{8pt plus 3pt minus 3pt}{1pt plus 1pt minus 1pt}

\definecolor{lime}{HTML}{A6CE39}
\DeclareRobustCommand{\orcidicon}{
	\begin{tikzpicture}
	\draw[lime, fill=lime] (0,0)
	circle [radius=0.16]
	node[white] {{\fontfamily{qag}\selectfont \tiny ID}};
	\draw[white, fill=white] (-0.0625,0.095)
	circle [radius=0.007];
	\end{tikzpicture}
	\hspace{-2mm}
}

\title{Deep Learning-Based Prediction of Suspension Dynamics Performance in Multi-Axle Vehicles}

\usepackage{authblk}

\author[1*]{
	Kai Chun Lin
	\href{https://orcid.org/0009-0003-1707-7653}{\orcidicon}
}

\affil[1]{Department of Computer Science, Rose-Hulman Institute of Technology}

\author[2*]{
	Bo-Yi Lin
	\href{https://orcid.org/0009-0002-8720-2774}{\orcidicon}
}

\affil[2]{Department of Mechanical Engineering, National Taiwan University}

\begin{document}

\twocolumn[\begin{@twocolumnfalse}

\maketitle

\begin{abstract}

This paper presents a deep learning-based framework for predicting the dynamic performance of suspension systems in multi-axle vehicles, emphasizing the integration of machine learning with traditional vehicle dynamics modeling. A Multi-Task Deep Belief Network Deep Neural Network (MTL-DBN-DNN) was developed to capture the relationships between key vehicle parameters and suspension performance metrics. The model was trained on data generated from numerical simulations and demonstrated superior prediction accuracy compared to conventional DNN models. A comprehensive sensitivity analysis was conducted to assess the impact of various vehicle and suspension parameters on dynamic suspension performance. Additionally, the Suspension Dynamic Performance Index (SDPI) was introduced as a holistic measure to quantify overall suspension performance, accounting for the combined effects of multiple parameters. The findings highlight the effectiveness of multitask learning in improving predictive models for complex vehicle systems.


\end{abstract}

\vspace{0.5cm}

\end{@twocolumnfalse}]


\section{Introduction}
Multi-axle vehicles are typically operated on rough and uneven terrain rather than on standard roads. The configuration of suspension system parameters has a significant impact on vehicle dynamics, and compared to two-axle vehicles, multi-axle vehicles are equipped with a greater number of suspension systems. This results in a wider range of possible parameter configurations. To maintain an acceptable level of ride comfort and vehicle handling on rough terrains, it is crucial to understand how different suspension system parameter configurations and vehicle-specific parameters affect vehicle dynamics. By doing so, methods can be developed to enhance the dynamic performance of the suspension system under such challenging conditions.

However, conducting physical prototyping and real-world testing presents high costs in terms of time and resources, particularly given the complexity and the presence of numerous unknown variables that can influence vehicle motion. These factors make experimental approaches to optimizing suspension system performance impractical. As a result, this study leverages a numerical model of a multi-axle half-vehicle to simplify the system’s complexity and to analyze the vertical and pitch motion of the vehicle. The primary objective of this research is to investigate the effects of varying vehicle parameters on the suspension dynamic performance of multi-axle vehicles.

Additionally, this study will develop a deep learning model to predict the influence of different vehicle parameters on the vehicle’s dynamic behavior. This predictive model aims to reduce the computational resources required for simulation, offering a more efficient and scalable method for evaluating suspension system performance in multi-axle vehicles. By integrating data-driven techniques with traditional numerical modeling, this research seeks to provide a novel framework for improving suspension dynamics and, consequently, the overall performance of multi-axle vehicles operating in harsh environments.

\section{Background} 

Vehicle dynamic analysis has been extensively studied using various vehicle models, with one of the most fundamental models being the quarter-car model \cite{Bastow1980CarSA, doi:https://doi.org/10.1002/9781119719984.ch7, reza_n__jazar_2009}. This simplified model is widely utilized to assess suspension system characteristics and performance. For instance, Chavan et al. \cite{Chavan2013ExperimentalVO} demonstrated that the simulation results of a quarter-car model align closely with experimental outcomes. Similarly, Prabhakar et al. \cite{Prabhakar2015SIMULATIONAA} utilized this model to examine the vibration isolation capabilities of suspension systems with varying spring and damping parameters under different road inputs.

Despite its utility, the quarter-car model’s limitations, such as its two-degree-of-freedom structure, necessitate more complex models to better capture the breadth of vehicle dynamics. The two-axle half-car model, featuring four degrees of freedom, provides a more comprehensive representation by incorporating both vertical and pitch motions of the vehicle. Researchers have applied this model to study various suspension performance metrics. For example, Faris et al. \cite{Faris2009AnalysisOS} used the half-car model to investigate the transfer function differences between passive and semi-active suspension systems in the frequency domain, while Vijayakumar et al. \cite{2002-01-0809} employed the same model to analyze its natural frequencies. In a related study, Rahman et al. \cite{Rahman2014InvestigationOV} extended the half-car model to five degrees of freedom by including the driver’s seat, thereby enabling an evaluation of ride comfort. Additionally, Soliman et al. \cite{2008-01-1146} utilized the half-car model to determine the vehicle parameters influencing ride comfort.

Research involving multi-axle vehicles and their suspension systems has been comparatively limited. Nonetheless, some studies have explored the dynamic performance of multi-axle systems. Faris et al. \cite{Faris2009AnalysisOS} applied a four-axle half-car model to examine the implementation of semi-active suspension systems. In another study, Ata \cite{Ata2014IntelligentCO} investigated control strategies for semi-active suspension systems in five-axle vehicles subjected to various road inputs, while El-Demerdash et al. \cite{2004-01-2079} analyzed multi-axle military vehicle performance on rough terrain. Sharaf et al. \cite{Mohsen2015ASA} compared multi-axle military vehicles with varying numbers of axles but identical vehicle parameters, assessing the effect of axle count on vehicle performance.

Most existing research has focused on evaluating suspension performance through key metrics such as ride comfort, suspension travel, and dynamic tire load. Studies on ride comfort \cite{Rauh2003VirtualDO, doi:10.1504/IJVD.1983.061313, HROVAT1988497} typically use sprung mass acceleration as the primary indicator. In contrast, research focusing on suspension travel \cite{HROVAT1988497, gillespie2021fundamentals, doi:10.1177/1077546305052315} emphasizes the limitations imposed by suspension design constraints. Meanwhile, dynamic tire load is commonly used as a measure of vehicle handling and control \cite{HROVAT1988497, gillespie2021fundamentals, Tseng1990SomeCO, NAGAYA2003477}.

In terms of vehicle performance optimization, the majority of research has concentrated on enhancing ride comfort. For example, Dong et al. \cite{10.1109/TCSVT.2024.3374758} proposed optimization techniques for suspension stiffness and damping in multi-axle off-road vehicles, focusing on improving ride smoothness. Additionally, Geweda et al. \cite{Geweda2017ImprovementOV} and Chen et al. \cite{Multi-objective} suggested multi-objective optimization strategies aimed at improving ride comfort.

Despite the extensive body of research on vehicle performance, spanning models from the quarter-car to multi-axle configurations, a notable gap remains in the literature concerning the dynamic performance of suspension systems in multi-axle vehicles. Most existing studies primarily focus on conventional metrics such as ride comfort, suspension travel, and dynamic tire load, with relatively little emphasis placed on the optimization of suspension dynamic performance. Moreover, few studies have systematically investigated the impact of variations in critical vehicle parameters on the dynamic behavior of suspension systems in multi-axle configurations.

Given the increasing complexity of vehicle systems and the nonlinear interactions between multiple parameters, conventional modeling and optimization approaches may struggle to capture the full scope of suspension dynamics in multi-axle vehicles. Recent advances in machine learning, particularly deep learning, have shown promise in modeling complex systems by identifying intricate patterns in large datasets. These methods provide a powerful framework for capturing the high-dimensional relationships inherent in multi-axle vehicle suspensions, enabling more accurate predictions of dynamic performance.

Motivated by these challenges and opportunities, this research aims to bridge the existing gap by leveraging deep learning techniques to predict the suspension dynamic performance of multi-axle vehicles. By integrating advanced data-driven approaches with classical vehicle dynamic models, this study seeks to develop a predictive framework that not only improves our understanding of suspension behavior but also facilitates the optimization of suspension systems in multi-axle vehicles. The outcomes of this research will offer novel insights into suspension design and performance enhancement, providing a foundation for future developments in vehicle systems.

\section{Methodology}

In this study, a comprehensive methodology is employed to analyze and predict the dynamic behavior and suspension performance of multi-axle vehicles under various conditions. The methodology integrates both numerical modeling and advanced machine learning techniques to provide a robust framework for evaluating suspension dynamics and optimizing vehicle design. The key components of this approach include the development of a multi-axle half-car model for dynamic analysis, the definition of performance metrics for suspension systems, and the application of a DBN-based multitask deep neural network for suspension dynamics performance prediction. Through this combined methodology, the study aims to deliver precise and insightful predictions that enhance vehicle handling and ride comfort across different road conditions.

\subsection{Multi-Axle Half-Car Model}
\label{sec:Multi-Axle Half-Car Model}

In order to analyze the dynamic characteristics of the vehicle, this study establishes a simplified numerical vehicle model based on fundamental assumptions aligned with the research objectives. To streamline the computational process and reduce complexity, the model simplifies the mass of various components of the vehicle into equivalent point masses. These are classified as the sprung mass and the unsprung masses corresponding to each axle. The sprung mass primarily consists of the vehicle body, payload, and powertrain, whereas the unsprung masses include the wheels, hubs, and suspension systems.

To investigate the dynamic behavior of the vehicle, the equations of motion for a multi-axle half-car model are formulated using Lagrange's equation, subject to a series of simplifying assumptions. Both the sprung and unsprung masses are modeled as rigid bodies, with their motion constrained to the vertical (Z) axis. The vehicle is assumed to travel at a constant velocity along the longitudinal (X) axis. The suspension system is represented as a linear system, with constant spring and damping coefficients. The tires are simplified as linear springs characterized by a stiffness coefficient, while the effects of tire rotational inertia are neglected. Additionally, the interaction between the tires and the road is modeled as point contact, and the road surface is assumed to be a rigid body. External forces, aside from gravitational forces and the normal forces exerted by the road on the tires, are disregarded in this model.

\begin{figure}
    \centering
    \includegraphics[width=0.95\linewidth]{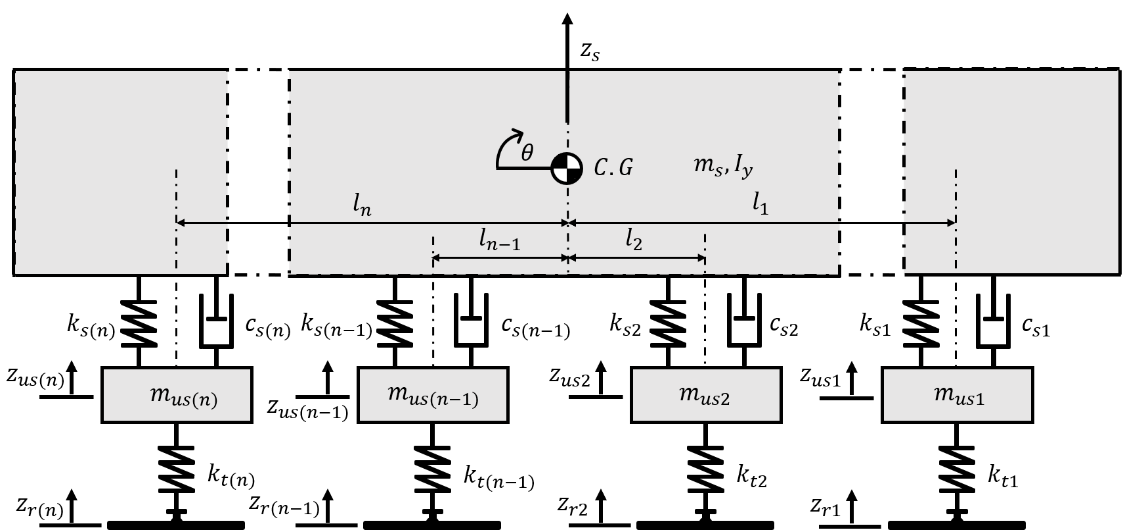}
    \caption{Multi-Axle Half-Car Model}
    \label{fig:Multi_Axle_Half_Car_Model}
\end{figure}

The multi-axle half-car model simplifies a multi-axle vehicle into a dynamic model with ${n+2}$ degrees of freedom representing the vertical bounce and pitch motion of the sprung mass, along with the vertical motion of the unsprung masses at each axle. As shown in Figure \ref{fig:Multi_Axle_Half_Car_Model}, the degrees of freedom correspond to the vertical and pitch movements of the sprung mass, and the vertical motion of the unsprung masses at each axle. This model serves as a tool to observe the dynamic behavior of vehicles with any number of axles.

After deriving the equations of motion, the matrix differential equations for the multi-axle half-car model can be expressed as:

\begin{equation}
    [M]\ddot{[z]} + [C]\dot{[z]} + [K][z] = [F]
\end{equation}

where the state vector \([z]\) is defined as:

\begin{equation}
    [z] = 
    \begin{bmatrix}
        z_s \\
        \theta \\
        z_{us(1)} \\
        \vdots \\
        z_{us(n)}
    \end{bmatrix}
\end{equation}

The mass matrix \([M]\) is given by:

\begin{equation}
    [M] = 
    \begin{bmatrix}
        m_s & 0 & 0 & \cdots & 0 \\
        0 & I_y & 0 & \cdots & 0 \\
        0 & 0 & m_{us(1)} & \cdots & 0 \\
        \vdots & \vdots & \vdots & \ddots & \vdots \\
        0 & 0 & 0 & \cdots & m_{us(n)}
    \end{bmatrix}
\end{equation}

The damping matrix \([C]\) is expressed as:

\begin{equation}
    \renewcommand{\arraystretch}{1.5}
    [C] = 
    \begin{small}
    \begin{bmatrix}
        \sum_{i=1}^{n} c_{s(i)} & \sum_{i=1}^{n} -l_{(i)}c_{s(i)} & -c_{s(1)} & \cdots & -c_{s(n)} \\
        \sum_{i=1}^{n} -l_{(i)}c_{s(i)} & \sum_{i=1}^{n} l^2_{(i)}c_{s(i)} & l_{(1)}c_{s(1)} & \cdots & l_{(n)}c_{s(n)} \\
        -c_{s(1)} & l_{(1)}c_{s(1)} & c_{s(1)} & \cdots & 0 \\
        \vdots & \vdots & \vdots & \ddots & \vdots \\
        -c_{s(n)} & l_{(n)}c_{s(n)} & 0 & \cdots & c_{s(n)}
    \end{bmatrix}
    \end{small}
\end{equation}

The stiffness matrix \([K]\) is defined as:

\begin{equation}   
    \renewcommand{\arraystretch}{1.5}
    [K] = 
    \begin{small}
    \begin{bmatrix}
        \sum_{i=1}^{n} k_{s(i)} & \sum_{i=1}^{n} -l_{(i)}k_{s(i)} & -k_{s(1)} & \cdots & -k_{s(n)} \\
        \sum_{i=1}^{n} -l_{(i)}k_{s(i)} & \sum_{i=1}^{n} l^2_{(i)}k_{s(i)} & l_{(1)}k_{s(1)} & \cdots & l_{(n)}k_{s(n)} \\
        -k_{s(1)} & l_{(1)}k_{s(1)} & k_{s(1)} + k_{t(1)} & \cdots & 0 \\
        \vdots & \vdots & \vdots & \ddots & \vdots \\
        -k_{s(n)} & l_{(n)}k_{s(n)} & 0 & \cdots & k_{s(n)} + k_{t(n)}
    \end{bmatrix}
    \end{small}
\end{equation}

Eventually, the external force vector \([F]\) is described as:

\begin{equation}
    [F] = 
    \begin{bmatrix}
        0 \\
        0 \\
        z_{r(1)}k_{t(1)} \\
        \vdots \\
        z_{r(n)}k_{t(n)}
    \end{bmatrix}
\end{equation}

\subsection{Suspension Dynamics Performance Definition}

The primary objective of this study is to analyze the dynamic performance of the suspension system in multi-axle vehicles. To achieve this, the dynamic behavior of the vehicle is simulated as it traverses a rough road surface, allowing for an examination of its motion characteristics and an analysis of the suspension's dynamic performance. To ensure that the simulated road closely resembles real-world conditions, this study employs a rough road input model based on the method proposed by Agostinacchio et al. \cite{Agostinacchio2014TheVI}, which generates random road profiles using the power spectral density of vertical displacement.

Building upon the work of Agostinacchio et al., this study defines the vehicle's suspension dynamic performance when traveling on a rough road using several key physical quantities. First, the sprung mass acceleration and angular acceleration are considered. As the values of these quantities decrease, the vehicle’s ride comfort improves. Thus, in this study, the root mean square (RMS) of the sprung mass acceleration is used as a key indicator of the suspension's dynamic performance, with lower RMS values representing better performance.

Second, the angular displacement of the sprung mass is analyzed. When the angular displacement of the sprung mass is reduced, the variations in normal forces on the wheels due to pitching motions are minimized, which enhances vehicle handling. Hence, the study defines the RMS of the sprung mass angular displacement as another measure of dynamic performance, with lower values indicating better handling.

Third, the suspension travel is considered. Smaller suspension travel implies that less space is required for suspension motion during vehicle operation. Consequently, this study defines the maximum suspension travel as an indicator of dynamic performance, where a smaller maximum value represents better performance.

Lastly, the dynamic tire load is assessed. This refers to the variation in the load on the tires relative to their equilibrium state. When the dynamic tire load is minimized, the variation in the tire's normal forces is reduced, resulting in greater effective lateral and longitudinal forces. This, in turn, improves the vehicle’s cornering, braking, and traction capabilities, leading to better handling. Thus, the RMS of the dynamic tire load is used as an important measure of suspension performance, with smaller values representing superior performance.

The study synthesizes these physical quantities to comprehensively define the suspension dynamic performance of the vehicle. According to the work of Cao \cite{Cao2008TheoreticalAO} and Griffin \cite{Qiu2005TransmissionOR}, it is noted that vehicle pitching movements are influenced more by driver control behavior than by the inherent properties of the vehicle. Therefore, the sprung mass acceleration, relative to angular acceleration and angular displacement, better reflects the comfort and handling characteristics attributable to the vehicle’s intrinsic properties. For multi-axle vehicles, handling is considered more important than ride comfort, making the dynamic tire load a more significant indicator of suspension dynamic performance than sprung mass acceleration, angular acceleration, or angular displacement. Meanwhile, suspension travel is merely a design constraint and is the least directly related to suspension performance.

Based on these considerations, this study concludes that dynamic tire load is the most critical indicator of suspension dynamic performance, followed by sprung mass acceleration, angular acceleration, and angular displacement. Suspension travel is the least indicative of performance. Consequently, a Suspension Dynamic Performance Index (SDPI) is defined to assess the suspension performance after adjustments to vehicle parameters when traveling on rough roads, as shown in the following equation:

\begin{align}
    SDPI = & \ 0.33 \cdot \left( 0.6 \cdot \frac{\ddot{{z}_s}_{RMS}}{\ddot{{z}_{s0}}_{RMS}} + 0.2 \cdot \frac{\ddot{\theta}_{RMS}}{\ddot{{\theta}_0}_{RMS}} + 0.2 \cdot \frac{\theta_{RMS}}{{\theta_0}_{RMS}} \right) \nonumber \\
    & + 0.01 \cdot \left( \frac{\sum_{i=1}^{n} {SWS_{(i)}}_{MAX}}{\sum_{i=1}^{n} {SWS_{0(i)}}_{MAX}} \right) \label{eq:SDPI} \\
    & + 0.66 \cdot \left( \frac{\sum_{i=1}^{n} {DTL_{(i)}}_{RMS}}{\sum_{i=1}^{n} {DTL_{0(i)}}_{RMS}} \right) \nonumber    
\end{align}

In Equation \ref{eq:SDPI}, each physical quantity for the vehicle after parameter adjustments is normalized by dividing it by the corresponding physical quantity before adjustments. This normalization ensures that physical quantities with different magnitudes and units can be compared. The weighting coefficients are determined based on the relationship between each physical quantity and suspension dynamic performance, ensuring that the SDPI of the vehicle prior to parameter adjustments is defined as 1. The smaller the SDPI, the better the suspension dynamic performance; conversely, the larger the SDPI, the worse the performance. Hence, the SDPI can be used to evaluate the impact of vehicle parameter changes on suspension dynamic performance.

\subsection{Numerical Model Simulation Parameters}

The parameters of the four-axle vehicle utilized in this study were derived from a combat vehicle, which served as the reference model. These parameters were then adjusted accordingly and subsequently employed to construct the numerical model. Furthermore, the assumptions applied in the simplified model are in alignment with those outlined in Section \ref{sec:Multi-Axle Half-Car Model}, ensuring consistency throughout the methodology. The final vehicle parameters, following these adjustments and considerations, are presented in Table \ref{tab:vehicle_table}.

\begin{table}
 \caption{Four-Axle Combat Vehicle Parameters}
  \centering
  \begin{tabular}{lllll}
    \toprule
    \multicolumn{2}{l}{\textbf{Name}} & \textbf{Symbol} & \textbf{Value} & \textbf{Unit} \\
    
    \midrule
    
    \multicolumn{2}{l}{Sprung mass}     & ${m_s}$         & 20337.8   & ${Kg}$  \\
    \multicolumn{2}{l}{Unsprung mass}   & ${m_{us}}$      & 458.4     & ${Kg}$   \\
    \multicolumn{2}{l}{Moment of inertia}  & ${I_y}$       & 562239.6  & ${Kg \cdot m^2}$   \\

    \multicolumn{2}{l}{Spring coefficient} & ${k_s}$ & 128710.0 & ${N/m}$\\
    \multicolumn{2}{l}{Damping coefficient} & ${c_s}$ & 11522.5  & ${N \cdot s/m}$\\
    \multicolumn{2}{l}{Tire stiffness coefficient}     & ${k_t}$ & 840857   & ${N/m}$\\
    \multicolumn{2}{l}{Wheelbase}      & ${wb}$    & 4.85    & ${m}$ \\

    \bottomrule
  \end{tabular}
  \label{tab:vehicle_table}
\end{table}

To compare the suspension dynamic performance across vehicle models with different numbers of axles, this study ensures that the tire load on each axle is identical among the models. As a result, the inter-axle distances are set equidistant, allowing the sprung mass to be evenly distributed across the axles. Additionally, the sprung mass of each model is adjusted according to the number of axles, with the pitch moment of inertia modified accordingly to reflect the changes in sprung mass.

\subsection{DBN-Based Multitask Deep Neural Network for Suspension Dynamics Performance Prediction}

Multitask learning (MTL) improves model performance by leveraging information from related tasks, enabling parallel learning where the knowledge acquired from each task enhances the learning process for others \cite{caruana1997multitask}. MTL addresses multiple related problems concurrently by employing a shared representation, which facilitates the partial sharing of input data across tasks. This approach capitalizes on commonalities between tasks, thereby enabling efficient knowledge transfer. Unlike conventional neural networks with multiple output layers, MTL models integrate task-specific features into a shared input feature vector, with hidden layers jointly used by all tasks. This structure proves particularly effective when training data is scarce.

Deep Belief Networks (DBNs) \cite{Hinton:2009}, composed of stacked layers of Restricted Boltzmann Machines (RBMs), are trained using a greedy layer-wise approach to extract hierarchical representations of input data. After each RBM layer is trained, its hidden layer output serves as the input for the subsequent layer.

\begin{figure}
    \centering
    \includegraphics[width=0.95\linewidth]{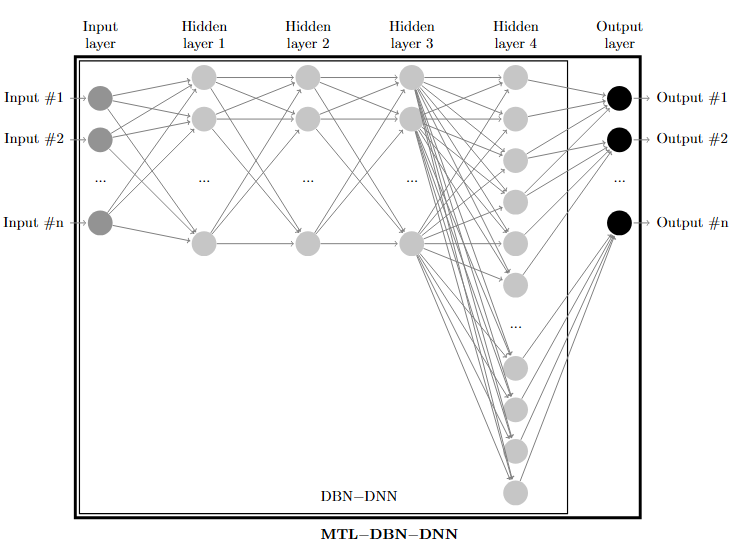}
    \caption{DBN-Based Deep Neural Network Model with Multitask Learning (MTL-DBN-DNN)}
    \label{fig:MTL_DBN_DNN}
\end{figure}

Li et al. \cite{https://doi.org/10.1155/2019/5304535} introduced a DBN-based multitask deep neural network (MTL-DBN-DNN) model, which addresses multiple tasks by utilizing shared information from the training data of various tasks. The architecture of this neural network is depicted in Figure \ref{fig:MTL_DBN_DNN}. The DBN component extracts high-level feature representations, while the output layer, utilizing a sigmoid activation function, manages the multitask predictions. Each output unit is connected to a specific subset of units in the final hidden layer of the DBN, with some units shared between adjacent subsets.

A notable advantage of this architecture lies in its locally connected layers, where only a subset of hidden units is assigned to each task. This is in contrast to fully connected networks, which must balance learning across all tasks and often fail to achieve optimal accuracy. By restricting each output unit’s connections to a subset of units in the preceding layer, the MTL-DBN-DNN model effectively captures task-specific information, as illustrated in Figure \ref{fig:MTL_DBN_DNN}.

\section{Results}
In this study, we utilized approximately 20000 data points generated through numerical simulations as input for training our MTL-DBN-DNN model. The model was trained over 50 epochs, with a batch size of 32. Through this process, we successfully reduced the average Mean Absolute Percentage Error (MAPE) to 0.026 and achieved an R² value of 0.98. Notably, the training and prediction phases took approximately 10 minutes, significantly outperforming traditional simulation methods, which typically require several hours to produce results. In addition, we conducted a comparative analysis between the MTL-DBN-DNN model and a conventional DNN model, demonstrating the superior efficiency and performance of the MTL-DBN-DNN approach. In addition, a sensitivity analysis was performed to evaluate the influence of various vehicle and suspension system parameters on dynamic suspension performance metrics.

\subsection{Comparative Analysis between the MTL-DBN-DNN Model and a Conventional DNN Model}

\begin{figure}
    \centering
    \includegraphics[width=0.95\linewidth]{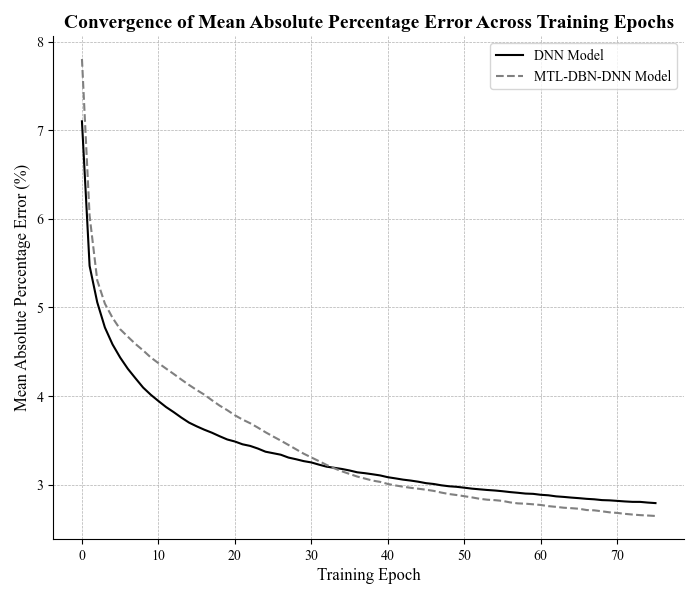}
    \caption{DBN-Based Deep Neural Network Model with Multitask Learning (MTL-DBN-DNN)}
    \label{fig:MAPE}
\end{figure}

To evaluate the performance of the multitask deep learning architecture, we conducted a comparative analysis between the MTL-DBN-DNN model and a conventional DNN model. Figure \ref{fig:MAPE} presents the convergence of the MAPE for both models over 70 training epochs.

As training progresses, both models show a decreasing trend in MAPE, indicating improving prediction accuracy. However, notable differences in their convergence behavior are observed. By the end of training, the MTL-DBN-DNN model achieves a significantly lower final MAPE compared to the conventional DNN, demonstrating superior overall performance.

These results suggest that the multitask deep learning architecture more effectively captures shared representations across tasks, leading to improved generalization. In contrast, the conventional DNN model, although showing stable convergence, lacks the benefits of multitask learning and consequently results in a relatively higher final MAPE.

Overall, these findings highlight the potential of multitask deep learning models in complex tasks such as suspension dynamics performance prediction, where related outputs can benefit from joint learning. The MTL-DBN-DNN model consistently outperforms traditional deep learning approaches by achieving lower error rates across multiple vehicle suspension performance metrics.

\subsection{Sensitivity Analysis of Vehicle and Suspension Parameters}

\begin{figure}
    \centering
    \includegraphics[width=1.0\linewidth]{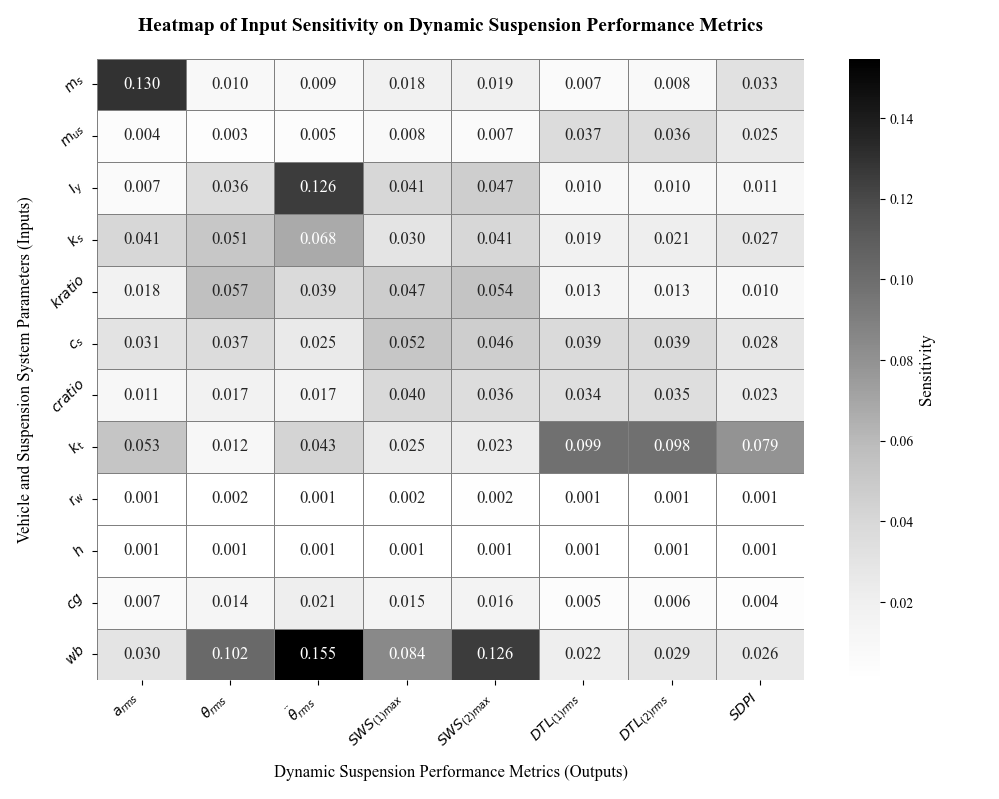}
    \caption{Heatmap of Input Sensitivity on Dynamic Suspension Performance Metrics}
    \label{fig:sensitivity_analysis}
\end{figure}

Figure \ref{fig:sensitivity_analysis} illustrates the sensitivity of various vehicle and suspension parameters on key dynamic suspension performance metrics. 

The analysis indicates that the sprung mass (\(m_s\)) exhibits the highest sensitivity with respect to sprung mass acceleration (\(a_{rms}\)), highlighting its crucial role in determining vehicle ride comfort and vertical dynamics. The pitch moment of inertia (\(I_y\)) shows a significant influence on pitch angular acceleration (\(\ddot{\theta}_{rms}\)), underscoring its critical importance in controlling pitch dynamics. Furthermore, both suspension stiffness (\(k_s\)) and damping coefficient (\(c_s\)) impact multiple performance metrics, playing a pivotal role in balancing ride comfort and handling capabilities. Tire stiffness (\(k_t\)) is found to notably influence dynamic tire load, making it a key factor in determining vehicle handling and stability. Finally, the wheelbase (\(wb\)) significantly affects suspension working space (\(SWS_{max}\)) and pitch angular acceleration (\(\ddot{\theta}_{rms}\)), emphasizing the critical role of vehicle geometry in overall suspension performance. 

These findings underscore the necessity of optimizing mass distribution, suspension properties, and vehicle geometry to improve dynamic suspension performance. The Suspension Dynamic Performance Index (SDPI) integrates these impacts, with parameters such as \(m_s\), \(k_s\), \(c_s\), \(k_t\), and \(wb\) contributing the most to performance variations.

\section{Conclusion}

In this study, we developed a deep learning-based approach for predicting suspension dynamics performance in multi-axle vehicles. The multi-task deep neural network (MTL-DBN-DNN) demonstrated significant improvements in prediction accuracy compared to traditional DNN models, as evidenced by the convergence of Mean Absolute Percentage Error (MAPE) across training epochs. The MTL-DBN-DNN model's ability to capture shared representations across multiple tasks proved to be highly effective, resulting in lower error rates and better generalization.

In addition, our sensitivity analysis revealed the critical impact of specific vehicle and suspension parameters on key dynamic suspension performance metrics. The Suspension Dynamic Performance Index (SDPI) successfully synthesized the influence of these parameters, offering a comprehensive measure for evaluating and optimizing suspension system performance.

The findings of this research underscore the potential of integrating data-driven approaches with traditional vehicle dynamics modeling to enhance the predictive accuracy and optimization of suspension systems. The MTL-DBN-DNN model's superior performance demonstrates its applicability to complex vehicular systems where multiple output variables benefit from joint learning.

\bibliography{reference}

\begin{thebibliography}{28}
\providecommand{\natexlab}[1]{#1}
\providecommand{\url}[1]{\texttt{#1}}
\expandafter\ifx\csname urlstyle\endcsname\relax
  \providecommand{\doi}[1]{doi: #1}\else
  \providecommand{\doi}{doi: \begingroup \urlstyle{rm}\Url}\fi

\bibitem[Bastow et~al.(1980)Bastow, Howard, and Whitehead]{Bastow1980CarSA}
Donald Bastow, Geoffrey Howard, and John~P. Whitehead.
\newblock \emph{Car suspension and handling}.
\newblock SAE International, 1980.

\bibitem[Wong(2022)]{doi:https://doi.org/10.1002/9781119719984.ch7}
J.Y. Wong.
\newblock \emph{Vehicle Ride Characteristics}, chapter~7, pages 469--520.
\newblock John Wiley and Sons, Ltd, 2022.
\newblock ISBN 9781119719984.
\newblock \doi{https://doi.org/10.1002/9781119719984.ch7}.

\bibitem[Jazar(2009)]{reza_n__jazar_2009}
Reza~N. Jazar.
\newblock \emph{Vehicle Dynamics: Theory and Application}.
\newblock Springer Cham, 2009.

\bibitem[Chavan et~al.(2013)Chavan, Sawant, and Tamboli]{Chavan2013ExperimentalVO}
Sayali~P Chavan, Sameer~H. Sawant, and Dipti~A. Tamboli.
\newblock Experimental verification of passive quarter car vehicle dynamic system subjected to harmonic road excitation with nonlinear parameters.
\newblock \emph{IOSR Journal of Mechanical and Civil Engineering}, 2013.

\bibitem[Prabhakar(2015)]{Prabhakar2015SIMULATIONAA}
Sairam Prabhakar.
\newblock Simulation and analysis of passive suspension system for different road profiles with variable damping and stiffness parameters.
\newblock \emph{Journal of Chemical and Pharmaceutical Sciences}, 2015.

\bibitem[Faris et~al.(2009)Faris, BenLahcene, and Ihsan]{Faris2009AnalysisOS}
Waleed~Fekry Faris, Zohir BenLahcene, and Sany~Izan Ihsan.
\newblock Analysis of semi-active suspension systems for four-axles off-road vehicle using half model.
\newblock \emph{International Journal of Vehicle Noise and Vibration}, 5:\penalty0 91--115, 2009.

\bibitem[{Vijayakumar, Srihari} and {Chandran, Ram S.}(2002)]{2002-01-0809}
{Vijayakumar, Srihari} and {Chandran, Ram S.}
\newblock Analysis of a 4-dof vehicle model using bond graph and lagrangian technique.
\newblock In \emph{SAE 2002 World Congress and Exhibition}. SAE International, mar 2002.
\newblock \doi{https://doi.org/10.4271/2002-01-0809}.

\bibitem[Rahman and Kibria(2014)]{Rahman2014InvestigationOV}
M~Shafiqur Rahman and Khan Muhammad~Golam Kibria.
\newblock Investigation of vibration and ride characteristics of a five degrees of freedom vehicle suspension system.
\newblock \emph{Procedia Engineering}, 90:\penalty0 96--102, 2014.

\bibitem[{Soliman, A. M. A.} et~al.(2008){Soliman, A. M. A.}, {Moustafa, S. M.}, and {Shogae, A. O. M.}]{2008-01-1146}
{Soliman, A. M. A.}, {Moustafa, S. M.}, and {Shogae, A. O. M.}
\newblock Parameters affecting vehicle ride comfort using half vehicle model.
\newblock In \emph{SAE World Congress and Exhibition}. SAE International, apr 2008.
\newblock \doi{https://doi.org/10.4271/2008-01-1146}.

\bibitem[Ata and Mohamed(2014)]{Ata2014IntelligentCO}
K.~I.~Mohammad Ata and Wael Mohamed.
\newblock Intelligent control of tracked vehicle suspension.
\newblock 2014.

\bibitem[{El-Demerdash, S. M.} and {Rabeih, E. M. A.}(2004)]{2004-01-2079}
{El-Demerdash, S. M.} and {Rabeih, E. M. A.}
\newblock Ride performance analysis of multi-axle combat vehicles.
\newblock In \emph{SAE 2004 Automotive Dynamics, Stability and Controls Conference and Exhibition}. SAE International, may 2004.
\newblock \doi{https://doi.org/10.4271/2004-01-2079}.

\bibitem[Mohsen et~al.(2015)Mohsen, Eltaher, Sharaf, and El-Demerdash]{Mohsen2015ASA}
M.~Mohsen, H.~M. Eltaher, A.~M. Sharaf, and Samir~M. El-Demerdash.
\newblock A systematic approach for the study and analysis of vehicle dynamics using design of experiments.
\newblock 2015.

\bibitem[Rauh(2003)]{Rauh2003VirtualDO}
Jochen Rauh.
\newblock Virtual development of ride and handling characteristics for advanced passenger cars.
\newblock \emph{Vehicle System Dynamics}, 40:\penalty0 135 -- 155, 2003.

\bibitem[Jolly(1983)]{doi:10.1504/IJVD.1983.061313}
A.~Jolly.
\newblock Study of ride comfort using a nonlinear mathematical model of a vehicle suspension.
\newblock \emph{International Journal of Vehicle Design}, 4\penalty0 (3):\penalty0 233--244, 1983.
\newblock \doi{10.1504/IJVD.1983.061313}.

\bibitem[Hrovat(1988)]{HROVAT1988497}
D.~Hrovat.
\newblock Influence of unsprung weight on vehicle ride quality.
\newblock \emph{Journal of Sound and Vibration}, 124\penalty0 (3):\penalty0 497--516, 1988.
\newblock ISSN 0022-460X.
\newblock \doi{https://doi.org/10.1016/S0022-460X(88)81391-9}.

\bibitem[Gillespie(2021)]{gillespie2021fundamentals}
T.~Gillespie.
\newblock \emph{Fundamentals of Vehicle Dynamics}.
\newblock SAE International, 2021.
\newblock ISBN 9781468601763.

\bibitem[Verros et~al.(2005)Verros, Natsiavas, and Papadimitriou]{doi:10.1177/1077546305052315}
G.~Verros, S.~Natsiavas, and C.~Papadimitriou.
\newblock Design optimization of quarter-car models with passive and semi-active suspensions under random road excitation.
\newblock \emph{Journal of Vibration and Control}, 11\penalty0 (5):\penalty0 581--606, 2005.
\newblock \doi{10.1177/1077546305052315}.

\bibitem[Tseng and Hrovat(1990)]{Tseng1990SomeCO}
T.~Tseng and Davorin~David Hrovat.
\newblock Some characteristics of optimal vehicle suspensions based on quarter-car models.
\newblock \emph{29th IEEE Conference on Decision and Control}, pages 2232--2237 vol.4, 1990.

\bibitem[Nagaya et~al.(2003)Nagaya, Wakao, and Abe]{NAGAYA2003477}
Go~Nagaya, Yasumichi Wakao, and Akihiko Abe.
\newblock Development of an in-wheel drive with advanced dynamic-damper mechanism.
\newblock \emph{JSAE Review}, 24\penalty0 (4):\penalty0 477--481, 2003.
\newblock ISSN 0389-4304.
\newblock \doi{https://doi.org/10.1016/S0389-4304(03)00077-8}.

\bibitem[He et~al.(2024)He, Xie, Xie, Jiang, and Chen]{10.1109/TCSVT.2024.3374758}
Lei He, Yongfang Xie, Shiwen Xie, Zhaohui Jiang, and Zhipeng Chen.
\newblock Iterative self-guided image filtering.
\newblock \emph{IEEE Trans. Cir. and Sys. for Video Technol.}, 34\penalty0 (8):\penalty0 7537–7549, mar 2024.
\newblock ISSN 1051-8215.
\newblock \doi{10.1109/TCSVT.2024.3374758}.

\bibitem[Geweda et~al.(2017)Geweda, El-Gohary, El-Nabawy, and Awad]{Geweda2017ImprovementOV}
A.~E. Geweda, Mohamed~A. El-Gohary, Amina E.~M. El-Nabawy, and Taher Awad.
\newblock Improvement of vehicle ride comfort using genetic algorithm optimization and pi controller.
\newblock \emph{alexandria engineering journal}, 56:\penalty0 405--414, 2017.

\bibitem[Liu et~al.(2024)Liu, Li, and Sa]{Multi-objective}
Z.~Liu, Z.~Li, and G.~Sa.
\newblock Multi-objective optimization of the vehicle ride comfort based on kriging approximate model and nsga-ii.
\newblock \emph{J Mech Sci Technol}, 38:\penalty0 4261–4276, 2024.
\newblock \doi{10.1007/s12206-024-0723-7}.

\bibitem[Agostinacchio et~al.(2014)Agostinacchio, Ciampa, and Olita]{Agostinacchio2014TheVI}
Michele Agostinacchio, Donato Ciampa, and Saverio Olita.
\newblock The vibrations induced by surface irregularities in road pavements – a matlab{\textregistered} approach.
\newblock \emph{European Transport Research Review}, 6:\penalty0 267--275, 2014.

\bibitem[Cao(2008)]{Cao2008TheoreticalAO}
Dongpu Cao.
\newblock Theoretical analyses of roll- and pitch-coupled hydro-pneumatic strut suspensions.
\newblock 2008.

\bibitem[Qiu and Griffin(2005)]{Qiu2005TransmissionOR}
Yi~Qiu and Michael~J. Griffin.
\newblock Transmission of roll, pitch and yaw vibration to the backrest of a seat supported on a non-rigid car floor.
\newblock \emph{Journal of Sound and Vibration}, 288:\penalty0 1197--1222, 2005.

\bibitem[Caruana(1997)]{caruana1997multitask}
Rich Caruana.
\newblock Multitask learning.
\newblock \emph{Machine learning}, 28:\penalty0 41--75, 1997.

\bibitem[Hinton(2009)]{Hinton:2009}
G.~E. Hinton.
\newblock {D}eep belief networks.
\newblock \emph{Scholarpedia}, 4\penalty0 (5):\penalty0 5947, 2009.
\newblock \doi{10.4249/scholarpedia.5947}.
\newblock revision \#91189.

\bibitem[Li et~al.(2019)Li, Shao, and Sun]{https://doi.org/10.1155/2019/5304535}
Jiangeng Li, Xingyang Shao, and Rihui Sun.
\newblock A dbn-based deep neural network model with multitask learning for online air quality prediction.
\newblock \emph{Journal of Control Science and Engineering}, 2019\penalty0 (1):\penalty0 5304535, 2019.
\newblock \doi{https://doi.org/10.1155/2019/5304535}.

\end{thebibliography}

\end{document}